\documentclass[11pt]{article}

\usepackage[preprint]{latex/acl}

\usepackage{times}
\usepackage{latexsym}
\usepackage{amsmath}
\usepackage[T1]{fontenc}
\usepackage[T5]{fontenc}

\usepackage[utf8]{inputenc}

\usepackage{microtype}

\usepackage{inconsolata}

\usepackage{graphicx}
\usepackage{booktabs}
\usepackage{multirow}
\title{DSC2025 – ViHallu Challenge: \\ Detecting Hallucination in Vietnamese LLMs}

\author{
  \textbf{Anh Thi-Hoang Nguyen}, 
  \textbf{Khanh Quoc Tran}, 
  \textbf{Tin Van Huynh}, \\
  \textbf{Phuoc Tan-Hoang Nguyen},
  \textbf{Cam Tan Nguyen},
  \textbf{Kiet Van Nguyen}\textsuperscript{1,2,3},
  \\[1.5ex] 
  \textsuperscript{1}Faculty of Information Science and Engineering,
  University of Information Technology\\
  \textsuperscript{2}University of Information Technology, Ho Chi Minh City, Vietnam\\
  \textsuperscript{3}Vietnam National University, Ho Chi Minh City, Vietnam 
  \\[1ex]
  \small{
    \textbf{Emails:} anhnth@uit.edu.vn
  }
}

\begin{document}
\maketitle
\begin{abstract}
The reliability of large language models (LLMs) in production environments remains significantly constrained by their propensity to generate hallucinations—fluent, plausible-sounding outputs that contradict or fabricate information. While hallucination detection has recently emerged as a priority in English-centric benchmarks, low-to-medium resource languages such as Vietnamese remain inadequately covered by standardized evaluation frameworks. This paper introduces the DSC2025 – ViHallu Challenge, the first large-scale shared task for detecting hallucinations in Vietnamese LLMs. We present the ViHallu dataset, comprising 10,000 annotated triplets of (context, prompt, response) samples systematically partitioned into three hallucination categories: no hallucination, intrinsic, and extrinsic hallucinations. The dataset incorporates three prompt types—factual, noisy, and adversarial—to stress-test model robustness. A total of 111 teams participated, with the best-performing system achieving a macro-F1 score of 84.80\%, compared to a baseline encoder-only score of 32.83\%, demonstrating that instruction-tuned LLMs with structured prompting and ensemble strategies substantially outperform generic architectures. However, the gap to perfect performance indicates that hallucination detection remains a challenging problem, particularly for intrinsic (contradiction-based) hallucinations. This work establishes a rigorous benchmark and explores a diverse range of detection methodologies, providing a foundation for future research into the trustworthiness and reliability of Vietnamese language AI systems.
\end{abstract}

\section{Introduction}
The widespread adoption of Large Language Models (LLMs) has fundamentally reshaped the landscape of Natural Language Processing (NLP) and Artificial Intelligence (AI). From automated content generation and machine translation to complex reasoning and conversational agents, LLMs have demonstrated capabilities that were, until recently, the exclusive domain of human cognition. However, as these models permeate critical sectors—ranging from healthcare and legal advising to customer support and education—a significant vulnerability has emerged as a primary bottleneck to their safe and reliable deployment: the phenomenon of hallucination.

Hallucination in LLMs refers to the generation of content that is fluent, grammatically coherent, and seemingly plausible, yet factually incorrect, nonsensical, or unfaithful to the provided source input. Unlike simple grammatical errors or obvious failures in fluency, hallucinations are insidious precisely because of their plausibility; a model may confidently assert a fabricated historical date, cite a non-existent legal precedent, or misinterpret a clear instruction while maintaining perfect syntax. This "illusion of competence" poses severe risks, particularly in high-stakes environments where misinformation can lead to financial loss, reputational damage, or safety hazards.

The challenge of mitigating hallucination is exacerbated in the context of low-to-medium resource languages. While English-centric models benefit from trillions of tokens of high-quality training data and extensive human feedback (RLHF), models for languages like Vietnamese often grapple with data scarcity, limited instruction-tuning datasets, and linguistic nuances that are not well-captured by multilingual architectures dominated by Indo-European data. Consequently, Vietnamese LLMs are often more susceptible to both intrinsic hallucinations (contradicting the input) and extrinsic hallucinations (fabricating information not present in the input).
In response to these urgent technical and safety challenges, the UIT Data Science Challenge 2025 (DSC 2025) introduces the ViHallu Challenge: Detecting Hallucination in Vietnamese LLMs. Organized as a Shared Task adhering to international standards set by venues such as SemEval and WMT, this competition aims to establish a robust, standardized benchmark for evaluating the reliability of Vietnamese LLMs. The challenge specifically addresses the need for automated systems capable of distinguishing between faithful generation and hallucinated content within the context of Question Answering (QA) and Dialogue tasks.

The development of LLMs has historically been driven by English-language benchmarks. Standard evaluation suites like MMLU, TruthfulQA, and HellaSwag have served as the north stars for model development. However, direct translation of these benchmarks into Vietnamese often fails to capture the cultural context, linguistic subtleties, and specific failure modes of models processing Vietnamese text. For instance, a model might perform exceptionally well on translated American history questions but fail catastrophically when asked about Vietnamese literature or legal codes, creating a false sense of reliability.

Furthermore, the nature of "hallucination" is context-dependent. In Retrieval-Augmented Generation (RAG) systems—a dominant architecture for enterprise AI—faithfulness to the retrieved context is paramount. If a RAG system answers a user's query by retrieving external knowledge trained into its weights rather than relying on the provided documents, it is technically hallucinating within the scope of that specific task, even if the information is factually correct in the real world. Existing generic benchmarks often conflate "world knowledge" with "contextual faithfulness," making them unsuitable for evaluating RAG reliability.

The ViHallu Challenge addresses this gap by providing a curated dataset that isolates the variable of faithfulness. By presenting models with a specific Context (Passage) and a Prompt, and asking participants to classify the Response as either faithful or hallucinated (and specifying the type), the challenge focuses on the model's reasoning and grounding capabilities rather than its memorized knowledge base.

\begin{figure}[h]
  \centering
  \includegraphics[width=\columnwidth]{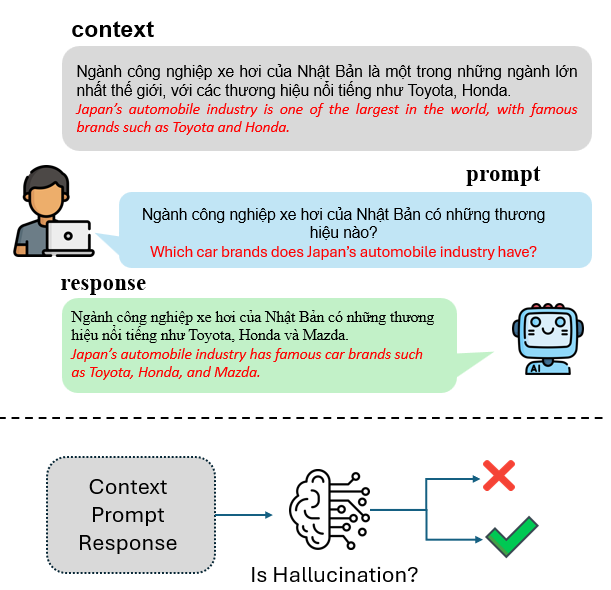}
  \caption{Illustration of the ViHallu task: given a context, prompt, and model response, the system predicts whether the response is hallucinated and, if so, its type.}
  \label{fig:task}
\end{figure}

The ViHallu Challenge is designed with three primary scientific and community goals:
\begin{enumerate}
    \item Benchmarking Reliability: To establish the first comprehensive, public benchmark for assessing the hallucination rates and robustness of Vietnamese LLMs. This moves the field away from anecdotal evidence of model failure toward rigorous, quantitative metrics.
    \item Advancing Mitigation Techniques: By framing hallucination detection as a classification task, the challenge encourages the development of novel architectures and methodologies. These may include advanced RAG verification loops, entailment-based classifier heads, uncertainty estimation via logit analysis, and post-editing mechanisms.
    \item Resource Democratization: The release of the ViHallu dataset under a CC-BY-SA 4.0 license provides the research community with a high-quality, human-annotated resource. This dataset will serve as a critical testbed for future research into AI safety for low-resource languages.
\end{enumerate}

In summary, the DSC 2025 ViHallu Challenge is not merely a competition but a strategic initiative to elevate the trustworthiness of Vietnamese AI. By rigorously defining the boundaries of hallucination and subjecting models to adversarial stress tests, the challenge aims to lay the groundwork for the safe integration of LLMs into the digital infrastructure of Vietnam.
\section{Background and Related Works}
The reliable deployment of Large Language Models (LLMs) is frequently hindered by the phenomenon of hallucination—instances where models generate plausible but factually incorrect or nonsensical information. To systematically address this challenge, the research community has converged on specific taxonomies to categorize these errors and established global benchmarks to measure them.

\subsection{Taxonomy of Hallucination: Intrinsic vs. Extrinsic}
The problem of hallucination in neural text generation has been studied extensively in the context of abstractive summarization, machine translation, and question answering. A foundational distinction, widely adopted across these domains, separates hallucinations into two primary categories: intrinsic and extrinsic~\cite{maynez-etal-2020-faithfulness}.

Intrinsic hallucinations occur when the model generates text that directly contradicts information present in the source context. These represent semantic or logical failures where the model manipulates available facts incorrectly, such as reversing entity relationships, swapping named entities, or drawing conclusions that refute the premises. For example, if a source describes "Company A acquiring Company B," an intrinsic hallucination might incorrectly state "Company B acquired Company A." Detecting intrinsic hallucinations fundamentally requires natural language inference (NLI) to determine whether the generated response entails a contradiction of the input context~\cite{10.1145/3703155}.

Extrinsic hallucinations, by contrast, involve the generation of information entirely absent from the source material. This category encompasses unverifiable claims, plausible-sounding fabrications, and factually correct statements that nonetheless violate the task requirement of faithfulness to the provided context. The critical distinction is that extrinsic hallucinations represent a failure of grounding rather than logical consistency. A model might correctly state that "Paris is the capital of France," but if this information does not appear in the source passage, it constitutes an extrinsic hallucination in a retrieval-augmented generation (RAG) or summarization setting where faithfulness to the input is paramount.

The ViHallu Challenge adopts this classical two-category schema with an additional "no hallucination" label for faithful responses. The challenge emphasises faithfulness to the provided context as the evaluation criterion, rather than factuality with respect to world knowledge. This framing aligns the task with practical deployment scenarios in which systems must be trustworthy and controllable, operating strictly within the information provided by users.

\subsection{Global Benchmarks and Shared Tasks}

The evaluation of hallucination detection in natural language generation has undergone significant evolution, transitioning from simple surface-level metrics to sophisticated, model-based benchmarks that assess semantic fidelity and factual consistency. Early work in this direction introduced TruthfulQA~\cite{lin-etal-2022-truthfulqa}, which measures whether language models generate answers that align with factual evidence or instead reproduce human misconceptions. This benchmark highlighted a critical insight: that fluent, grammatically correct outputs can nonetheless harbour subtle factual errors, motivating the development of more targeted detection methodologies.

Parallel efforts in fact extraction and verification, exemplified by the FEVER series~\cite{thorne-etal-2018-fever}, established the foundations for claim-level verification pipelines. FEVER operates by extracting atomic claims from text and matching them against a knowledge base, a process that has evolved in recent iterations such as AVeriTeC to incorporate real-world web search and handle more open-domain claims. These benchmarks demonstrate the feasibility of explicit verification procedures and have influenced the design of retrieval-augmented approaches to hallucination mitigation.

At the task competition level, the SemEval shared-task series has recently elevated hallucination detection to a primary focus. SHROOM 2024 (Task 6)~\cite{mickus-etal-2024-semeval} introduced a systematic evaluation of fluent but incorrect outputs across machine translation, paraphrase generation, and definition modeling, offering both model-aware (with access to generator internals) and model-agnostic (black-box) evaluation protocols. The follow-up Mu-SHROOM 2025 (Task 3)~\cite{vazquez-etal-2025-semeval} significantly expanded the scope to fourteen languages and shifted from sentence-to-span-level hallucination identification, requiring systems to identify the exact character offsets of erroneous regions. However, despite this multilingual expansion, Vietnamese was not included in Mu-SHROOM's coverage, highlighting a gap in large-scale hallucination evaluation for Southeast Asian languages.

Parallel to international shared tasks, the Conference on Machine Translation (WMT) Quality Estimation (QE) tasks have explicitly targeted critical error detection, particularly hallucinations in machine translation. WMT 2024 QE~\cite{zerva-etal-2024-findings} introduced sub-tasks for identifying hallucinations—instances where translations introduce content unrelated to source material—and incorporated automatic post-editing challenges, encouraging systems to move beyond detection toward error correction. The ViHallu Challenge adopts this philosophy by optionally accepting corrected response proposals, inviting teams to explore mitigation strategies alongside detection.

At the large-scale LLM level, HaluEval~\cite{li-etal-2023-halueval} represents a methodological breakthrough, employing a ``sampling-then-filtering'' pipeline in which instruction-tuned models are prompted to generate hallucinations intentionally, which are subsequently filtered and verified by human annotators. Providing 35,000 samples across QA, dialogue, and summarization tasks, HaluEval demonstrates how to scale hallucination benchmark construction through generative sampling paired with rigorous human verification.

Overall, the emergence of multilingual, fine-grained, and LLM-focused hallucination detection tasks reflects the field's recognition that reliability and faithfulness are prerequisite properties for deploying language models in high-stakes applications. The ViHallu Challenge positioned within this landscape, addresses a critical gap by providing the first large-scale, community-driven benchmark for Vietnamese LLM hallucination detection.
\section{The DSC 2025 - LLM Hallucination Challenge}
The DSC 2025 – ViHallu Challenge establishes a standardized evaluation campaign to rigorously assess the reliability of Large Language Models in Vietnamese. Moving beyond traditional accuracy metrics, this challenge quantifies how well models resist generating false information (hallucination) and their stability under imperfect input conditions (robustness). This section details the task definition, the dataset structure, the evaluation methodology, and the operational timeline.

\subsection{Task Definition}
The objective of the challenge is to classify the faithfulness of a Vietnamese LLM's response given a specific context and prompt. Each sample in the dataset consists of:

\begin{enumerate}
    \item \textbf{Context ($C$):} A reference passage (1--3 sentences).
    \item \textbf{Prompt ($P$):} A question or instruction. Prompts are categorized as \textit{Factual} (standard), \textit{Noisy} (containing typos/errors), or \textit{Adversarial} (containing distractors).
    \item \textbf{Response ($R$):} The answer generated by the target LLM.
\end{enumerate}

Participants must build a system to predict a label $L \in \{\text{no}, \text{intrinsic}, \text{extrinsic}\}$:
\begin{itemize}
    \item \textbf{No Hallucination (\texttt{no}):} The response is fully consistent with the information in the passage. It does not contain any unsupported information. It correctly answers the prompt based only on the provided context.

    \item \textbf{Intrinsic Hallucination (\texttt{intrinsic}):} The response contradicts or distorts information specifically mentioned in the passage. The model misinterprets entities, numbers, or relationships present in the source.

    \item \textbf{Extrinsic Hallucination (\texttt{extrinsic}):} The response contains additional information not found in the passage. Crucially, even if the information is factually true in the real world (e.g., general knowledge), if it cannot be derived from the passage, it is classified as extrinsic. 
\end{itemize}

\subsection{Evaluation Metrics}
The official ranking metric is \textbf{Macro-F1}, chosen to ensure balanced performance across all three classes regardless of class distribution:

\begin{equation}
    \text{Macro-F1} = \frac{1}{3} \sum_{c \in \{\text{no, intrinsic, extrinsic}\}} F1_c
\end{equation}

\noindent \textbf{Accuracy} is used as a secondary metric to break ties in the leaderboard.
\subsection{Schedule and Overview Summary}
The DSC2025 – ViHallu Challenge followed a structured timeline designed to provide sufficient development time for participants while maintaining competitive momentum. Table \ref{tab:schedule} summarizes the key dates and phases of the competition.

\begin{table}[h]
\caption{Schedule of the DSC 2025 – ViHallu Challenge}
\label{tab:schedule}
\resizebox{\columnwidth}{!}{%
\begin{tabular}{ll}
\hline
\textbf{Times} & \textbf{Phase}                                                                    \\ \hline
August 15th    & Registration Period                                                               \\
August 29th    & Warm-up Phase                                                                     \\
September 5th  & Public Test Phase                                                                 \\
October 3rd    & Private Test Phase                                                                \\
October 5th    & Competition end                                                                   \\ \hline
October 7th    & Submission deadline                                                               \\
October 24th   & \begin{tabular}[c]{@{}l@{}}Results Announcement \\ \& Award Ceremony\end{tabular} \\ \hline
\end{tabular}%
}
\end{table}

The competition schedule spanned approximately 10 weeks, with the core development phase lasting 28 days (September 5 – October 3) on the public test set, providing substantial opportunity for iterative model refinement and experimentation. The three-day private test phase (October 3–5) enforced submission limits (3 per day) to prevent overfitting while enabling final system tuning.

\begin{table}[h]
\caption{Participation Summary of the DSC 2025 – ViHallu Challenge}
\label{tab:participation_summary}
\resizebox{\columnwidth}{!}{%
\begin{tabular}{ll}
\hline
\textbf{Metric}                     & \textbf{Value} \\ \hline
\#Registration Teams                & 155            \\
\#Teams with Signed Data Agreements & 136            \\
\#Submitted Teams                   & 111            \\
\#Paper Submissions                 & 3              \\ \hline
\end{tabular}%
}
\end{table}

The DSC2025 – ViHallu Challenge attracted significant community interest, with 155 teams registering from universities and organizations across Vietnam and internationally. Of these, 136 teams (87.7\%) proceeded to sign data use agreements and access the training dataset, indicating strong commitment to the shared task. During the public test phase, 111 teams (71.6\% of registered teams) made active submissions to the leaderboard, representing a solid participation rate comparable to international shared tasks.

The competition culminated in 3 system papers submitted by top-performing teams, documenting their technical approaches and findings. Following standard shared task practice, participating teams were encouraged but not required to submit system papers describing their methods. Top teams received recognition and opportunities to present their work at the DSC2025 awards ceremony.
\section{Corpus Creation}
The development of the UIT-ViHallu corpus followed a rigorous, multi-step methodology designed to ensure high fidelity and reproducibility. Adhering to established practices for shared task dataset construction, the process evolved from initial data collection through iterative annotation and cross-checking, concluding with a final validation phase. This section details the complete pipeline used to generate the benchmark dataset for the DSC2025 – ViHallu Challenge.
\subsection{Data Construction Overview}
As illustrated in Figure 2, the corpus creation workflow is divided into four primary stages: (i) Data Collection, (ii) Data Annotation, (iii) Cross-checking, and (iv) Validation. This structured approach was implemented to maintain systematic quality control at every step, ensuring that the final dataset serves as a reliable benchmark for evaluating hallucination and robustness in Vietnamese LLMs.
\begin{itemize}
    \item \textbf{Stage 1: Data Collection}
    \paragraph{Passage and Prompt-Response Pool Generation} The foundational passages were sourced from UIT-ViQuAD 2.0 \cite{Nguyen_2022}, a prominent Vietnamese machine reading comprehension dataset comprising over 35,000 question-answer pairs derived from Wikipedia. We employed stratified random sampling to select approximately 10,000 passages, assigning each a unique identifier (passage\_id). The selected texts range from 88 to 1500 tokens in length, a span chosen to reflect the realistic document snippets typically encountered in retrieval-augmented generation (RAG) and document-grounded dialogue systems.

    For every selected passage, we generated three distinct prompt configurations to test different aspects of model performance:
    \begin{itemize}
        \item \textit{Factual prompts}: These are direct, grammatically standard questions derived from the passage content using extraction-based templates. They cover entity, relationship, and temporal queries intended to establish a baseline for model performance.
        \item \textit{Noisy prompts}: To assess robustness against input noise, we applied systematic perturbations to factual prompts. These included diacritic removal—a critical challenge in Vietnamese processing—along with character swaps, token deletions, and word reordering. Crucially, these perturbations were controlled to ensure the prompt remained semantically interpretable to human readers.
        \item \textit{Adversarial prompts}: These prompts were synthesized using an LLM to contain misleading presuppositions, false premises, "trap" logic, or entailment reversals. The objective was to actively induce hallucinations by challenging the model's ability to adhere to the source context.
    \end{itemize}

    \paragraph{Response Generation} For each resulting (passage, prompt) triple, responses were produced using a state-of-the-art instruction-following LLM (GPT-4o) executed under controlled and deterministic decoding settings. The model was prompted with a standardized instruction template enforcing strict grounding to the provided passage. A single response was collected per instance to capture the model’s natural behavior across factual, noisy, and adversarial prompt conditions.

    To ensure quality and consistency, we conducted human validation on a randomly sampled 10\% subset of all generated outputs. Annotators verified adherence to grounding constraints, coherence of Vietnamese text, and absence of extraneous information not supported by the passage. This sampling-based inspection served as an additional safeguard against systematic generation errors.
    
    To further assess whether any artifacts were tied to a specific model, we performed a cross-model comparison on a smaller subset of prompts. The same instruction template was executed using GPT-4o, GPT-4o-mini, and GPT-4.1-mini, enabling us to identify potential model-specific biases or hallucination patterns. The comparison confirmed that the distributional characteristics observed in the responses were consistent across model variants.
    
    All finalized triples were serialized into JSONL format, with complete metadata including model configuration, decoding parameters, and timestamps. This stage yielded a raw dataset of approximately 10,000 instances with a balanced distribution across the three prompt types.
    
    \item \textbf{Stage 2: Data Annotation} 
    \paragraph{Annotation Setup and Guidelines} We recruited a team of 12 annotators, all of whom possess formal training in Natural Language Processing and native-level proficiency in Vietnamese. Prior to the main task, the team underwent comprehensive training based on the finalized annotation protocols (Guideline). This training covered operational definitions for the three target labels (no, intrinsic, extrinsic), decision rules for handling borderline or ambiguous cases, and a review of 12 worked examples (spanning all label and prompt types). The team also utilized a checklist procedure to ensure systematic label assignment.
    
    \paragraph{Annotation Process} The annotation was conducted in batches of approximately 970 samples to facilitate workload management and continuous quality monitoring. For each (passage, prompt, response) triple, annotators independently assigned a hallucination label and an optional confidence score on a 1–5 scale. For ambiguous cases, annotators were required to provide written justifications. A key instruction was to assess the response's faithfulness strictly relative to the provided passage, rather than its real-world factual correctness. This constraint ensures the dataset aligns with the practical requirements of RAG systems, where grounding in retrieved documents is paramount.

    \item \textbf{Stage 3: Cross-checking} 
    \paragraph{Peer Review and Conflict Resolution} To guarantee data reliability, we implemented a systematic cross-checking phase adapted from standard multilingual shared task practices. In this phase, each annotator independently reviewed a random subset (10–15\%) of annotations contributed by their peers. Reviewers assessed the correctness of the assigned labels, the accuracy of citations used as supporting evidence, and the presence of any syntax errors or typos in the text fields. Any instance where the reviewer disagreed with the initial annotation was flagged for further resolution.
    
    \paragraph{Disagreement Resolution Protocol}
    Discrepancies identified during the review phase were handled through a structured three-step protocol. First, disagreement cases were analyzed to identify recurring patterns of confusion, such as the boundary between intrinsic and extrinsic hallucination or thresholds for inferability. Second, a senior annotator with extensive experience in hallucination detection acted as an adjudicator, reviewing the justifications from both the original annotator and the reviewer to cast a tie-breaking vote. Finally, all adjudication decisions were logged with detailed rationales, creating an internal knowledge base of edge cases. This process ensured bidirectional consistency, verifying that subtle hallucinations were identified uniformly across the dataset.

    \item \textbf{Stage 4: Validation}
    \paragraph{Validation Criteria and Automated Checks} The finalized dataset underwent a dual-layer validation process involving both automated scripts and manual verification. Automated checks ensured technical integrity, verifying JSON structural validity, the uniqueness of IDs across all fields, and the completeness of required fields (passage, prompt, response, label). The scripts also enforced text encoding standardization (UTF-8) and checked that response lengths fell within the acceptable range of 5–250 tokens.
    
    \paragraph{Manual Quality Assessment and Exclusion} Concurrently, supervisors performed manual validation by spot-checking 5\% of random samples per batch. This assessment focused on the consistency between labels and evidence, adherence to Guideline, and linguistic quality. Samples were removed if they failed specific criteria, such as inconsistent labeling (e.g., applying "extrinsic" to a direct contradiction), malformed text, or temporal inconsistencies.
\end{itemize}

\subsection{Overview Statistics of UIT-ViHallu}
This section analyzes the UIT-ViHallu corpus, focusing on label distribution, text length characteristics, and lexical patterns across the training and test partitions.

\paragraph{Dataset Composition}
The final corpus consists of 10,000 samples divided into three stratified subsets: a training set (70\%), a public test set (10\%), and a private test set (20\%). These samples were derived from 4,351 unique passages, indicating that multiple prompt-response pairs were generated from single source passages. This one-to-many mapping maximizes data diversity while optimizing the manual annotation effort.

\paragraph{Label Distribution}
Table~\ref{tab:label_dist} details the class distribution across the dataset partitions. The training set maintains a balanced distribution, with intrinsic hallucinations being slightly more frequent (34.97\%), followed closely by extrinsic hallucinations (32.96\%) and faithful responses (32.07\%). This balance is preserved in the test sets; for instance, the private test set shows a nearly even split ($\approx$ 32--34\% per class).

This uniform distribution is significant for evaluation validity. Unlike datasets where the ``faithful'' class dominates, ViHallu forces detection systems to distinguish between valid and hallucinated content without relying on majority-class priors. The data suggests that under the specific prompting conditions of this task (factual, noisy, and adversarial), the baseline model (GPT-4o) produced faithful, contradictory, and hallucinatory responses at comparable rates.

\begin{table}[h]
    \centering
    \resizebox{\columnwidth}{!}{%
    \begin{tabular}{lcccc}
        \toprule
        \textbf{Dataset} & \textbf{No Hallucination} & \textbf{Intrinsic} & \textbf{Extrinsic} & \textbf{Total} \\
        \midrule
        Train & 2,245 & 2,448 & 2,307  & 7,000 \\
        Public Test & 334 & 344 & 322 & 1,000 \\
        Private Test & 690 & 672 & 638 & 2,000 \\
        \midrule
        \textbf{Overall} & \textbf{3,269} & \textbf{3,464} & \textbf{3,267} & \textbf{10,000} \\
        \bottomrule
    \end{tabular}%
    }
    \caption{Label distribution across dataset partitions.}
    \label{tab:label_dist}
\end{table}

\paragraph{Text Length Statistics}
We analyzed the token counts for passages (contexts), prompts, and responses using whitespace tokenization to ensure consistent, language-agnostic measurement. As summarized in Table~\ref{tab:length_stats}, the length statistics are highly consistent across the three splits, confirming the effectiveness of the stratified sampling strategy.

\begin{itemize}
    \item \textbf{Contexts:} Passages average approximately 180 tokens but exhibit high variance (std $\approx$ 72), with lengths ranging from 88 to over 1,500 tokens. This variation reflects the diverse nature of the source Wikipedia articles, requiring models to handle both short summaries and longer, detailed documents.
    \item \textbf{Prompts:} Prompts average roughly 27 tokens. The distribution captures the spectrum of difficulty, from concise factual queries (e.g., ``DNA là gì?'' -- 3 tokens) to complex adversarial prompts involving multiple clauses (up to 94 tokens).
    \item \textbf{Responses:} Model responses show a stable mean length of approximately 40 tokens across all partitions. The low standard deviation ($\approx$ 10) indicates that the generator model maintained a consistent level of verbosity regardless of the input type.
\end{itemize}

\begin{table}[h]
    \centering
    \setlength{\tabcolsep}{3pt} 
    \resizebox{\columnwidth}{!}{
    \begin{tabular}{llcccc}
        \toprule
        \textbf{Metric} & \textbf{Dataset} & \textbf{Mean} & \textbf{Min} & \textbf{Max} & \textbf{Std} \\
        \midrule
        \multirow{3}{*}{\textit{Context Length}} 
            & Train        & 179.7 & 88 & 1,537 & 72.5 \\
            & Public Test  & 179.4 & 88 & 671   & 68.9 \\
            & Private Test & 177.5 & 88 & 1,537 & 71.7 \\
        \midrule
        \multirow{3}{*}{\textit{Prompt Length}} 
            & Train        & 26.8 & 3 & 94 & 13.4 \\
            & Public Test  & 26.8 & 3 & 73 & 13.0 \\
            & Private Test & 26.6 & 4 & 82 & 12.9 \\
        \midrule
        \multirow{3}{*}{\textit{Response Length}} 
            & Train        & 39.5 & 1 & 68 & 10.3 \\
            & Public Test  & 39.8 & 5 & 66 & 9.6 \\
            & Private Test & 39.6 & 1 & 64 & 10.0 \\
        \bottomrule
    \end{tabular}%
    }
    \caption{Statistics of text length of UIT-ViHallu dataset}
    \label{tab:length_stats}
\end{table}

\begin{figure*}[h]
    \centering
     \includegraphics[width=\linewidth]{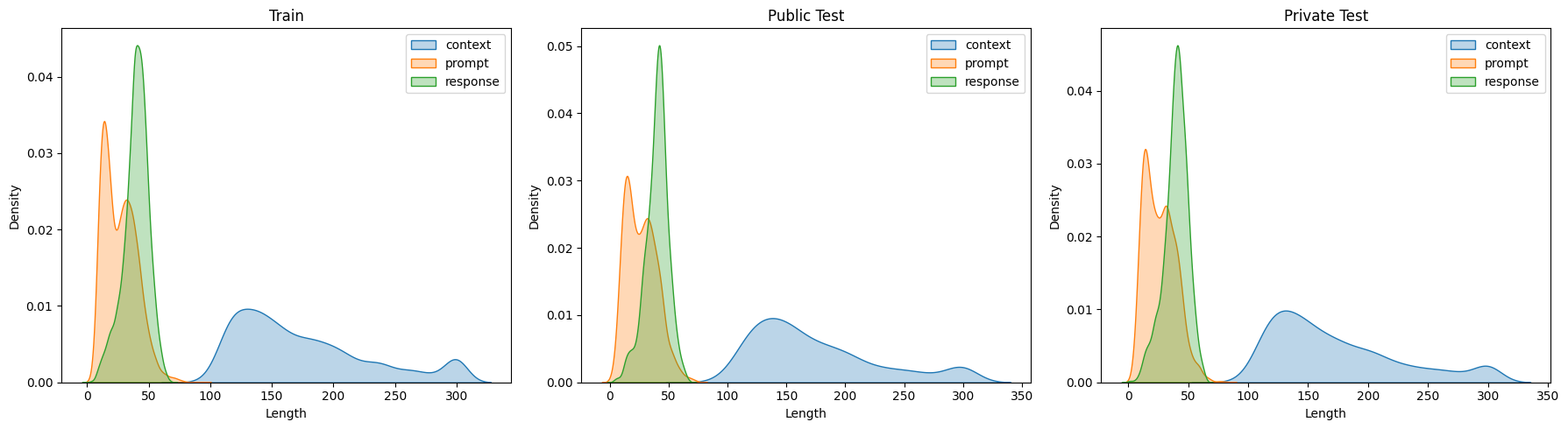}
     \caption{Kernel density estimates of token lengths for Context, Prompt, and Response across dataset splits.}
     \label{fig:length_dist}
\end{figure*}

\paragraph{Length Distribution Analysis}
Figure~\ref{fig:length_dist} illustrates the kernel density estimates for token lengths. The distributions for the training, public test, and private test sets overlap almost perfectly, validating that the test sets are representative of the training data distribution.

Three distinct patterns emerge from the density plots:
\begin{enumerate}
    \item \textbf{Passages} follow a right-skewed unimodal distribution, peaking between 130--160 tokens, consistent with standard paragraph lengths in encyclopedic text.
    \item \textbf{Prompts} display a bimodal distribution. The primary peak around 15--20 tokens corresponds to standard factual questions, while the secondary peak at 30--40 tokens reflects the longer, more complex structure of adversarial and noisy prompts.
    \item \textbf{Responses} exhibit a near-Gaussian distribution centered at 40 tokens, suggesting that while the prompt types varied significantly, the length of the model's output remained relatively constrained.
\end{enumerate}

\section{System and Results}
\subsection{Baseline System}
To establish a performance floor, a baseline system was constructed using a standard multilingual encoder PhoBERT \cite{nguyen-tuan-nguyen-2020-phobert}, fine-tuned as a three-way classifier. Each input triplet was serialized into a single sequence in the form
\([CLS]\) context \([SEP]\) prompt \([SEP]\) response \([SEP]\),
and the \([CLS]\) representation was fed to a linear classification head. The model was trained with cross-entropy loss and the AdamW optimizer, without any task-specific architectural modifications or prompt engineering. This simple configuration yields only modest performance, with macro-F1 around 0.30 and accuracy close to 0.33 on the development and private test sets, and thus mainly serves as a lower bound for comparison.

\subsection{Challenge Submissions}
The shared task attracted 111 submissions. The final leaderboard shows a clear trend toward instruction-tuned Large Language Models (LLMs) in the 4–7B parameter range, often combined with parameter-efficient fine-tuning and ensemble strategies. At the same time, one competitive system relies on strong encoder-based NLI models integrated through stacking.

\subsubsection{\textit{The HCMUS-ThangQuang Team}}
HCMUS-ThangQuang deployed a single Qwen3-4B-Instruct model adapted with LoRA. Training used an effective batch size of 32, a learning rate of \(5 \times 10^{-5}\), and one epoch over the training data. The key design choice is a structured prompt that first describes the task and the three hallucination labels, then provides several illustrative examples, and finally presents the instance to classify. This explicit instruction format helps the model reason about consistency between context and response. The system ranks first on the private test set with a macro-F1 of 84.80\%.

\subsubsection{\textit{The HCMUTransformer Team}}
HCMUTransformer adopted a more elaborate ensemble-based approach. They fine-tuned 35 LoRA adapters on top of Qwen3-Embedding-4B, varying the input preprocessing across adapters: some receive raw text, others use Vietnamese spelling correction, and a third group applies both correction and similarity-based context reduction. The final prediction is obtained by a weighted combination of adapter outputs, where the non-negative weights are learned via Sequential Least Squares Programming to maximise macro-F1 on a validation split. This method places second on the private test set with 84.73\% macro-F1 and also performs strongly on the public test (83.16\%).

\subsubsection{\textit{The UIT\_WhiteCow Team}}
UIT\_WhiteCow built a dual-LLM system using Qwen3-4B and Gemma3-4B. Both models are fine-tuned with LoRA and few-shot prompting. At inference time, each model is run with ten different temperature values, from 0.0 to 0.9, and the labels produced across all temperatures and both models are aggregated by majority vote. This temperature-based voting scheme is intended to stabilise predictions under ambiguous or adversarial prompts. Although the team does not appear in the top part of the public leaderboard, their system reaches 84.54\% macro-F1 on the private test, ranking third overall.

\subsubsection{\textit{The UIT\_Champion Team}}
UIT\_Champion focused on high-capacity encoder models. They fine-tuned both Vietnamese and English NLI transformers, including DeBERTa-xlarge-MNLI and RoBERTa-large-MNLI, and translated the ViHallu data into English to exploit existing NLI pretraining. The outputs of all fine-tuned encoders are then combined by an XGBoost meta-classifier that operates on the predicted class probabilities. This stacking-based approach achieves 84.19\% macro-F1 on the private test set.

\subsubsection{\textit{The 3MoTB Team}}
3MoTB proposed a three-stage architecture centred on Qwen models. First, three Qwen variants (Qwen3-4B, Qwen2.5-7B, and Qwen2.5-7B-Instruct) are fine-tuned with LoRA (rank 16). Second, the class probability vectors from these models are concatenated and passed through a small feed-forward neural network to obtain an initial ensemble prediction. Third, for instances where this ensemble disagrees with Qwen2.5-7B, the system invokes a set of binary “expert resolvers” based on Qwen3-4B, each specialised on one pair of labels. These experts decide between the competing labels. The system obtains 84.42\% macro-F1 on the private test set and also appears in the top part of the public leaderboard (82.83\%).

\subsubsection{\textit{The UIT-HalluGuard Team}}
UIT-HalluGuard combined multiple LLaMA and Qwen families at different precision levels. After preliminary experiments with encoder models for context selection, they fine-tuned several LLaMA and Qwen variants using LoRA, including a chain-of-thought version of Qwen3-4B. Their final solution follows a two-stage majority-voting scheme: predictions are first aggregated within each model family (LLaMA, Qwen3-4B, Qwen2.5-7B), then the three family-level decisions are combined by a second majority vote. This hierarchical ensemble reaches 84.23\% macro-F1 on the private test set.

\subsubsection{\textit{The Prime Team}}
Prime addressed the task via multi-step binary decision making. Instead of directly predicting one of three labels, they trained three binary classifiers: one to detect the presence of any hallucination, one to decide whether all information in the response is grounded in the context, and one to identify contradictions. A rule-based module combines these binary outputs into the final three-way label, while a separate three-class model acts as a backup when the rules do not apply. After analysing typical errors, the team added a small post-processing step to correct a few extrinsic hallucinations that had been mislabelled as faithful responses. The final system reaches 84.01\% macro-F1 on the private test set.

\subsection{Results}
\begin{table*}[h]
    \centering
    \caption{Performance of baseline and top-7 systems on the public and private test sets.}
    \label{tab:leaderboard}
    \resizebox{\textwidth}{!}{%
    \begin{tabular}{cllcc}
        \toprule
        \textbf{Rank} & \textbf{Team} & \textbf{Key Techniques} & \textbf{Public Test F1} & \textbf{Private Test F1} \\
        \midrule
        1 & HCMUS-ThangQuang & Single Qwen3-4B, structured prompting & 0.8295 & 0.8480 \\
        2 & HCMUTransformer   & 35 LoRA adapters with SLSQP-weighted ensemble & 0.8316 & 0.8473 \\
        3 & UIT\_WhiteCow     & Qwen3 + Gemma3 with temperature-based voting & --     & 0.8454 \\
        4 & 3MoTB             & Qwen ensemble plus expert resolvers          & 0.8283 & 0.8442 \\
        5 & UIT-HalluGuard    & LLaMA + Qwen in two-phase majority voting    & --     & 0.8423 \\
        6 & UIT\_Champion     & Vietnamese/English NLI encoders + XGBoost stacking & -- & 0.8419 \\
        7 & Prime             & Binary decomposition with rule-based fusion  & --     & 0.8401 \\
        \midrule
        -- & BASELINE         & Encoder-only (PhoBERT)                      & 0.3228 & 0.3283 \\
        \bottomrule
    \end{tabular}%
    }
    \begin{flushleft}
        \footnotesize \textit{Note: Public test scores are shown only for teams that appeared in the organiser’s public leaderboard snapshot.}
    \end{flushleft}
\end{table*}

As shown in Table~\ref{tab:leaderboard}, all top systems outperform the encoder-only baseline by a large margin. While the baseline remains around 0.33 macro-F1 on the private test set, every system in the top seven exceeds 0.84 macro-F1, demonstrating the importance of instruction-tuned models, ensemble strategies, and task-aware prompting for hallucination detection in Vietnamese. At the same time, the best private-test score of 0.8480 indicates that the task remains challenging, leaving room for further improvements in modelling contextual contradictions and unsupported information.

\section{Conclusion and Future Work}
The DSC2025 – ViHallu Challenge establishes the first comprehensive benchmark for hallucination detection in Vietnamese LLMs. With 111 participating teams and a dataset of 10,000 annotated samples, the challenge demonstrates that hallucination detection is tractable yet non-trivial. The best-performing system achieves 84.80\% macro-F1, representing a 51-point improvement over the encoder-only baseline, while highlighting persistent challenges in distinguishing intrinsic hallucinations from faithful responses.

Key findings reveal that instruction-tuned LLMs with structured prompting outperform complex ensemble architectures, that architectural diversity enables multiple viable solutions, and that the problem remains unsolved despite strong empirical results. The public release of the ViHallu dataset under CC-BY-SA 4.0 provides a valuable resource for the research community.

Future work should focus on retrieval-augmented verification for extrinsic detection, contrastive learning for intrinsic hallucinations, confidence calibration for safer deployment, and span-level annotations for finer-grained analysis. Extending the benchmark to other Southeast Asian languages would broaden its impact and support the development of trustworthy multilingual AI systems.
\section*{Acknowledgments}

\bibliography{latex/custom}




\end{document}